\RequirePackage[T1]{fontenc}
\documentclass[letterpaper,10pt,conference]{ieeeconf}
\IEEEoverridecommandlockouts
\usepackage{amsmath,amssymb,graphicx,booktabs}
\usepackage{newtxtext,newtxmath}
\usepackage{cite}
\usepackage{array}
\title{\LARGE\bf From Documented Strengths to Force Limits: Material-Informed Robotic Insertion for Construction Assembly}
\author{Lin He$^{1}$, Yanyi Chen$^{1}$, Haofei Sun$^{2}$, Lingyao Li$^{3}$, and Min Deng$^{1*}$%
\thanks{$^{*}$Corresponding author: Min Deng.}%
\thanks{$^{1}$Lin He, Yanyi Chen, and Min Deng are with the Department of Civil and Environmental Engineering,
The University of Tennessee, Knoxville, TN 37996, USA
(e-mail: \{lynnhe,yanychen,mindeng\}@utk.edu).}%
\thanks{$^{2}$Haofei Sun is with The University of Texas at Arlington, Arlington, TX 76019, USA
(e-mail: hxs7410@mavs.uta.edu).}%
\thanks{$^{3}$Lingyao Li is with The University of Arizona, Tucson, AZ 85721, USA
(e-mail: lingyaoli@arizona.edu).}%
}
\begin{document}
\bstctlcite{IEEEexample:BSTcontrol}
\maketitle
\thispagestyle{empty}
\pagestyle{empty}

\begin{abstract}

Insertion is a fundamental operation in robotic construction assembly, where variations in material properties and assembly conditions make it difficult to select contact forces that complete the task without exceeding the assembly's capacity. Although construction documents encode engineering knowledge about materials and their conditions, translating this knowledge into load limits for a specific assembly remains difficult. This paper presents SAGE (Source-grounded Assembly Gating and Execution), a system that converts documented material evidence into capacity estimates for robotic insertion. SAGE restricts a large language model (LLM) to extracting tensile and compressive strengths from retrieved passages and tables and records their sources. A response model then interpolates offline finite element (FE) solutions to convert these strengths and the assembly conditions into axial load capacity. For fits with positive clearance, the estimated capacity sets the policy's axial force limit; for interference fits, it is compared with measured support demand to determine admission. On the primary benchmark, SAGE reduces mean capacity error from 80.65\% for direct LLM estimates based on the same evidence to 10.74\%. Without refitting, the mean error remains 8.00\% on 16 additional geometries. Under the assigned support release model, SAGE correctly classifies 59 of 62 scored simulation runs, with only conservative errors. In recorded xArm6 demonstrations, SAGE takes material documents as input and completes physical insertion in 9 of 13 trials. These results show that assigning document interpretation to the LLM and force calculation to an explicit mechanical model produces accurate capacity estimates and traceable insertion decisions.

\end{abstract}

\section{INTRODUCTION}

Robotic construction assembly offers the potential to improve precision, alleviate labor demands, and reduce worker exposure to hazardous environments~\cite{brosque2022impacts,chen2026perception}, and insertion is a fundamental operation in this setting~\cite{bock2015automation}. However, reliable robotic insertion remains challenging because construction tasks vary substantially in material properties, component thickness, support conditions, and interface friction~\cite{izard2023pressfits}. These variations affect both the force required for insertion and the load that the assembly can sustain, making it difficult to coordinate motion, force, and compliance within the assembly's mechanical limits~\cite{hogan1985impedance}.

To address this control challenge, prior work has developed a force-conditioned reinforcement learning policy that adapts insertion behavior to a prescribed axial force limit over a continuous range~\cite{priorpolicy}. The policy seeks efficient insertion while respecting the prescribed axial force limit, but it takes this limit as given and does not determine an appropriate value for a particular material, material state, or assembly.

In construction, product documentation and technical literature provide information on materials and their condition~\cite{aac}. However, they report material grades and test strengths rather than force limits for a specific assembly. Estimating the assembly's load capacity requires identifying the relevant properties and converting them into loads. For example, compression and splitting tests for autoclaved aerated concrete (AAC) report different quantities~\cite{aacpores}, while wood strength depends on direction and moisture~\cite{wood}. LLMs are well suited to this setting because they can interpret heterogeneous technical sources and translate natural-language construction instructions into robot actions~\cite{mechgpt,he2025llm,chen2026perception}. Retrieval can ground material properties in relevant passages and tables~\cite{rag}, but it does not provide the mechanical conversion required for a specific assembly. In our benchmark, direct LLM capacity estimates based on the same retrieved evidence have a mean error of 80.65\%. SAGE therefore retrieves relevant passages and tables~\cite{rag}, restricts the LLM to extracting cited material properties, and uses a fixed response model built from offline FE solutions to convert those properties into assembly capacity. The resulting force limit is grounded in both documentary evidence and an explicit mechanical calculation.

Capacity also affects execution differently across fit conditions. With sufficient clearance between the hole and peg, alignment can avoid sustained wall contact, whereas misalignment can cause jamming~\cite{whitney1982quasistatic}. Small-clearance assembly, in which the hole is only marginally larger than the peg, is commonly studied in force-aware insertion research~\cite{forge}. Interference fits introduce another practically important loading condition: because the hole is smaller than the peg, insertion produces sustained radial contact pressure even after alignment~\cite{gai2021compliance}. These contact loads can induce damaging stresses that are not reflected in the axial insertion force alone. This condition motivates a separate admission check that compares the capacity with the support demand.

We propose SAGE (Source-grounded Assembly Gating and Execution), a system that estimates assembly capacity from material documents and uses the estimates to guide robotic insertion. SAGE restricts the LLM to extracting cited material strengths, converts them into assembly capacity using a fixed FE response model, and uses the estimate to set the insertion policy input or make an admission decision. Our contributions are as follows.
\begin{enumerate}
\item \textbf{Source-grounded material extraction.} An LLM extracts tensile and compressive strengths with supporting passages from retrieved text and tables. On the primary benchmark, linked retrieval increases valid capacity coverage by 30 percentage points over basic extraction.

\item \textbf{Mechanical capacity estimation.} We develop an FE response model that converts the extracted strengths into assembly capacity. Against refined FE references, the mean error is 10.74\% on the initial geometries and 8.00\% on additional geometries without refitting.

\item \textbf{Capacity-guided insertion.} We transfer the estimated capacity to an insertion policy and evaluate an additional admission check for interference fits. SAGE correctly classifies 59 of 62 scored simulation runs with only conservative errors, while recorded xArm6 demonstrations complete insertion for two materials.
\end{enumerate}

\section{RELATED WORK}

\subsection{Force-Aware Robotic Insertion}
Classical analyses relate peg insertion forces to geometry, friction, and jamming~\cite{whitney1982quasistatic}. Learning-based methods improve insertion under contact uncertainty through residual control~\cite{johannink2019residual}, visuo-haptic representations that generalize across geometries and clearances~\cite{lee2019visiontouch}, and variable impedance policies that adjust pose and compliance during contact~\cite{martin2019vices}. Most closely related, FORGE conditions an insertion policy on a prescribed force limit and uses force feedback and dynamics randomization during execution~\cite{forge}. These methods determine how the robot should insert once a force limit is given, but not how that limit follows from the material strength and geometry of the assembly. SAGE supplies this input while leaving the insertion policy unchanged.

\subsection{Language Models for Robot Decisions}
Language models select feasible robot skills by combining language scores with learned skill values~\cite{saycan}, flag uncertain decisions through conformal prediction~\cite{ren2023knowno}, and convert language and visual observations into spatial value maps~\cite{huang2023voxposer}. In construction, site updates can revise planning constraints and coordinate robot tasks through a digital twin~\cite{deng2025llm_digital_twin}. These systems address task feasibility and coordination rather than material load capacity. FORGE-plus is closer, using object descriptions and force feedback to select recovery actions, but its force limits remain assigned inputs~\cite{forgeplus}. SAGE instead derives the constraint used by the robot before execution.

\subsection{From Material Evidence to Robot Constraints}
Retrieval augmented generation supplies supporting passages to a language model~\cite{rag}, and dense passage retrieval learns semantic representations for retrieving relevant evidence~\cite{karpukhin2020dpr}, but retrieval alone does not convert a reported material strength into a load for a specific geometry and contact condition. Engineering language models go further: MechGPT organizes mechanics knowledge from technical sources~\cite{mechgpt}, MechAgents generates and revises finite element code~\cite{mechagents}, and Wilke combines material inference with computational assessment~\cite{wilke}. Solver access alone, however, does not provide a fixed and traceable conversion from documentary evidence to a robot constraint. SAGE separates this process into cited property extraction, deterministic capacity estimation, and capacity-guided insertion.

\section{METHOD}
\begin{figure*}[t]
\centering
\includegraphics[width=\textwidth]{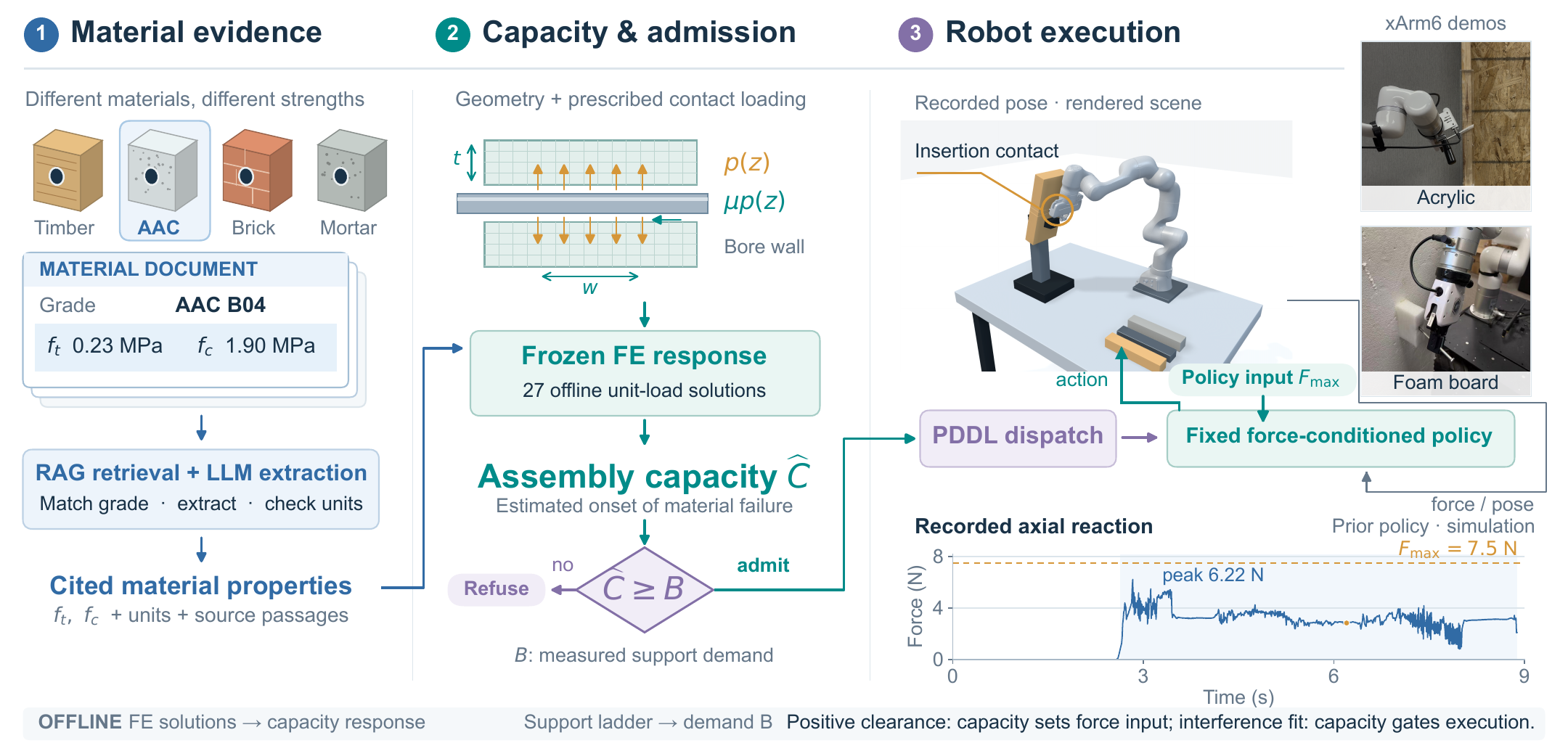}
\caption{Overview of SAGE. Documented strengths are converted to capacity under prescribed loading. The estimate sets the policy's axial force input for positive clearance or is compared with support demand before an interference fit is dispatched. The workcell and force trace show the same recorded simulation of the prior insertion policy~\cite{priorpolicy}; these workcell visuals are illustrative. The photographs are recorded xArm6 hardware runs.}
\label{fig:method}
\end{figure*}

Fig.~\ref{fig:method} outlines how SAGE connects material documentation to robotic insertion through three stages. First, an LLM extracts tensile and compressive strengths from retrieved passages and tables and records their sources. Second, a fixed response model interpolates offline finite element (FE) solutions to estimate the axial load at which a prescribed stress criterion is first reached under the specified contact loading. Third, SAGE transfers this estimate to execution: it serves as the policy's axial force input for insertions with positive clearance, while an admission gate compares it with measured support demand for interference fits. A Planning Domain Definition Language (PDDL)~\cite{ghallab1998pddl} interface dispatches tasks admitted by the gate to the pretrained insertion policy. The photographs show xArm6 hardware demonstrations with acrylic and foam-board assemblies.

\subsection{Problem Formulation}

We define the assembly capacity $C_{\mathrm{ax}}$ as the axial load at which a prescribed material stress criterion is first reached under the specified contact loading. Because this capacity affects execution differently depending on whether wall contact persists after alignment, we consider two fit conditions: positive clearance and interference. Our primary setting has a small positive clearance, meaning that the hole is slightly larger than the peg, and uses a pretrained insertion policy $\pi(a_t \mid o_t,F_{\max})$~\cite{priorpolicy} conditioned on a prescribed axial force limit. Applying this policy to different materials and assemblies requires selecting an $F_{\max}$. In the interference condition, the hole is smaller than the peg, so insertion produces persistent contact pressure. Limiting the axial force alone therefore does not determine whether the resulting local stresses are compatible with the material strength.

The inputs are material documents $\mathcal{D}$, material identity and state $m$, assembly geometry $g$, and prescribed loading and support conditions $\ell$. Because the documents provide material strengths rather than $C_{\mathrm{ax}}$, our objective is to estimate $\widehat{C}_{\mathrm{ax}}$ from $(\mathcal{D},m,g,\ell)$.

\subsection{Material Evidence Extraction}
\label{sec:extraction}
Estimating $C_{\mathrm{ax}}$ requires material strengths applicable to the specified material and state. Each retrieval query uses the material grade, state, loading direction, and requested strength statistic. Paragraphs and tables are indexed separately and ranked by cosine similarity between their TF-IDF representations. The retriever returns eight passages and follows explicit table references to include up to four additional tables, preserving numerical data with their explanatory context.

An LLM extracts tensile and compressive strengths from the retrieved evidence. Each output includes the value, unit, availability flag, and a quotation linked to its source passage. The prompt restricts the LLM to property extraction rather than capacity or $F_{\max}$ prediction, leaving the conversion from material strength to load to the mechanical model.

The extracted records then undergo numerical and evidence checks. Numerical validation accepts finite, positive values in supported units and converts them to pascals. Evidence validation checks whether each quotation appears in its cited passage after whitespace normalization. These checks support traceability but do not verify that the reported test or material condition is applicable. We denote the resulting tensile and compressive strengths by $f_t$ and $f_c$, respectively, and use them as inputs to the capacity model described next.

\subsection{Capacity Estimation and Force Limit Transfer}
\label{sec:capacity}
\label{sec:force_selection}

Using the extracted strengths $f_t$ and $f_c$, we estimate the axial capacity of the assembly under prescribed contact loading. The material surrounding the hole is modeled as a homogeneous isotropic annulus with bore radius $a$, wall thickness $t_{\mathrm{w}}$, and height $H$. Its bottom surface is constrained axially but remains free radially, while the other exterior surfaces are traction free. The bore geometry and nominal elastic properties are fixed across the response library, with their values and rationale reported in the experimental setup. Because the linear elastic problem is traction controlled, Young's modulus scales displacement but does not affect the stress coefficients used below; fixing Poisson's ratio is an explicit modeling assumption.

The bore is loaded by a smooth pressure distribution over an axial span $w$:
\begin{equation}
p(z)=
\begin{cases}
P\sin^2\!\left(\dfrac{\pi(z-z_0)}{w}\right),
& z_0<z<z_0+w,\\
0, & \text{otherwise},
\end{cases}
\label{eq:pressure}
\end{equation}
where $P$ is the pressure amplitude. The loaded span is centered at $z=18$~mm, giving $z_0=18~\mathrm{mm}-w/2$. Assuming fully mobilized sliding friction with coefficient $\mu$, the axial traction is $-\mu p(z)$. Its integral over the bore gives the axial load magnitude
\begin{equation}
F(P)
=2\pi a\mu\int_{z_0}^{z_0+w}p(z)\,\mathrm{d}z
=\mu\pi a wP.
\label{eq:axial_load}
\end{equation}

Under small-strain linear elasticity, stress scales with $P$. We use axisymmetric bilinear quadrilateral elements with $2\times2$ Gauss quadrature and evaluate the principal stresses at the element Gauss points. Let $c_t$ denote the peak tensile principal stress per unit pressure amplitude, and let $c_c$ denote the magnitude of the most compressive principal stress per unit pressure amplitude. The pressure at which either material strength is first reached is
\begin{equation}
P_{\mathrm{crit}}
=\min\left(\frac{f_t}{c_t},\frac{f_c}{c_c}\right).
\label{eq:critical_pressure}
\end{equation}
Substitution into Eq.~\eqref{eq:axial_load} gives
\begin{equation}
C_{\mathrm{ax}}
=\mu\pi a w
\min\left(\frac{f_t}{c_t},\frac{f_c}{c_c}\right).
\label{eq:capacity}
\end{equation}

The resulting $C_{\mathrm{ax}}$ denotes the onset of the prescribed stress criterion, rather than collapse after cracking. Using splitting tensile strength as an isotropic tensile threshold is an modeling assumption. The pressure and friction distributions are also prescribed rather than obtained from evolving insertion contact, so the estimate applies only when these loading assumptions represent the assembly being considered.

To avoid an FE solve for every query, we construct an offline response library over a grid of wall thickness $t_{\mathrm{w}}$, loading span $w$, and friction coefficient $\mu$. A unit-pressure solve at each grid point provides the coefficients $c_t$ and $c_c$. A query $(t_{\mathrm{w}},w,\mu)$ is located in the grid cell enclosing it and rescaled to normalized coordinates $\alpha,\beta,\gamma\in[0,1]$, one per axis. Each coefficient is then the weighted sum of that cell's eight corner values,
\begin{equation}
\widehat{c}=\sum_{i,j,k\in\{0,1\}}\alpha_i\beta_j\gamma_k\,c^{(i,j,k)},\quad
\alpha_1=\alpha,\ \alpha_0=1-\alpha,
\label{eq:trilinear}
\end{equation}
with $\beta_j$ and $\gamma_k$ analogous. Applying Eq.~\eqref{eq:trilinear} to the tensile and compressive coefficients gives $\widehat{c}_t$ and $\widehat{c}_c$, yielding
\begin{equation}
\widehat{C}_{\mathrm{ax}}
=\mu\pi a w
\min\left(
\frac{f_t}{\widehat{c}_t},
\frac{f_c}{\widehat{c}_c}
\right).
\label{eq:estimated_capacity}
\end{equation}
Geometry and friction determine the interpolated response, while $f_t$ and $f_c$ determine the material thresholds. The library remains fixed across material queries, and queries outside its domain are rejected.

For insertions with positive clearance, we transfer the estimate to the policy when the prescribed loading represents the contact loading produced during execution:
\begin{equation}
F_{\max}=\operatorname{clip}\!\left(
\widehat{C}_{\mathrm{ax}},F_{\mathrm{lo}},F_{\mathrm{hi}}
\right),
\label{eq:force_selection}
\end{equation}
where $\operatorname{clip}$ restricts its first argument to $[F_{\mathrm{lo}},F_{\mathrm{hi}}]$, with $F_{\mathrm{lo}}=3$~N and $F_{\mathrm{hi}}=11$~N the bounds used for the policy input in this work. The mapping reads only $\widehat{C}_{\mathrm{ax}}$: the refined FE capacity $C^\star$ is computed to score estimation accuracy in Sec.~\ref{sec:capacity_accuracy}, is not available at execution time, and therefore never enters the transferred force limit. It supplies an input within the policy's operating range but does not guarantee an instantaneous force bound or simulate material damage. If $\widehat{C}_{\mathrm{ax}}<F_{\mathrm{lo}}$, an application that must remain below the estimate should refuse execution rather than clip the input upward.

\subsection{Additional Admission Check for Interference Fits}
\label{sec:admission}

The preceding mapping from capacity to the policy input applies to insertions with positive clearance. Interference fits need an additional check because the axial force limit no longer bounds the load that matters: the hole is smaller than the peg, so radial contact pressure persists after alignment and sets the local stresses. These stresses are also hard to monitor during execution, because opposing radial forces largely cancel in the resultant support reaction while the material remains stressed. Whether to attempt the insertion must therefore be decided before it starts.

The gate leaves the prescribed axial force limit unchanged and compares the capacity estimate with the support demand of the insertion. In simulation, the support is a dynamic body held to the world by a breakable PhysX fixed joint with release threshold $T$: when the constraint force at the joint exceeds $T$, the joint breaks and the support detaches, which we call release. For a fixed interface and test condition, we sweep $T$ over increasing values---a support ladder---and take the empirical demand $B$ as the lowest value at which the support holds throughout execution. Thus, $B$ represents the executor's support requirement rather than the material strength or axial force limit.

To evaluate admission, we assign the refined FE reference capacity $C^\star$ to $T$ for each case while holding the sleeve geometry and contact configuration fixed. Under this assigned correspondence, the gate is
\begin{equation}
\text{admit}\iff\widehat{C}_{\mathrm{ax}}\ge B.
\label{eq:gate}
\end{equation}
Missing or invalid estimates lead to refusal, while quotation mismatches are recorded for traceability but do not independently trigger refusal.

Assigning $T=C^\star$ provides a common release model for measuring how capacity estimation errors affect admission. It does not equate $C^\star$ with a measured support reaction or independently validate material failure. Release indicates that the assigned threshold $T$ was exceeded, rather than simulated cracking, and the estimated capacity remains a point estimate rather than a calibrated design allowable. Tasks that pass the gate retain their prescribed axial force limit and proceed to policy execution.

\subsection{Insertion Policy Execution}
\label{sec:execution}
Once the force limit is specified and any required admission check is passed, the insertion action is dispatched with $F_{\max}$ to the pretrained policy~\cite{priorpolicy}, which controls motion and impedance during insertion. A PDDL interface~\cite{ghallab1998pddl} coordinates peg pickup, alignment, and insertion, using admission as a precondition for insertion. The policy weights and execution configuration remain unchanged across comparisons. The precondition is enforced in normal operation; for admission evaluation only, we bypass it so that refused cases also yield a recorded outcome to score the gate against.

\section{EXPERIMENTS}
\begin{figure*}[!t]
\centering
\includegraphics[width=\textwidth,trim=0 14 0 2,clip]{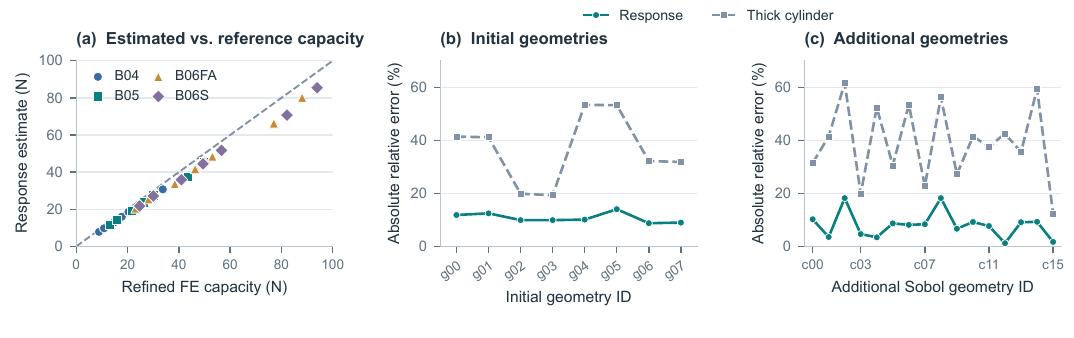}
\caption{Capacity estimation with known material strengths. (a) Response estimates versus refined FE capacities for four AAC grades at eight initial geometries; the diagonal denotes exact agreement. (b) Absolute relative errors of the response model and thick cylinder approximation for initial geometries g00--g07. (c) Errors for additional Sobol geometries c00--c15.}
\label{fig:capacity}
\end{figure*}

\subsection{Evaluation Protocol}

The capacity benchmark uses four AAC grades~\cite{aacpores} and brick MU15~\cite{brick} at eight geometries excluded from response construction. Without refitting, interpolation is also tested using the four AAC grades at sixteen geometries within the response domain. Extraction diagnostics include additional documents and material conditions, and LLM evaluations use two fixed generation seeds. Simulation robot experiments use the pretrained policy~\cite{priorpolicy}; force limit transfer uses the original sleeve, while admission evaluation also uses a stiff sleeve with higher support demand.

The response library, retrieval and extraction settings, force mapping, admission rule, FE convergence criterion, and policy configuration are fixed before evaluation. Refined FE capacities $C^\star$ are used only to compute error and are never supplied to the estimation methods. Mechanical comparisons use identical geometries, loading conditions, stress criteria, and strengths; document comparisons also use identical retrieved evidence. Quantitative force-transfer and admission comparisons use simulation with matched initial conditions, while the xArm6 experiments evaluate physical execution and are reported separately.

\subsection{Capacity Estimation Accuracy}
\label{sec:capacity_accuracy}

To isolate the mechanical conversion from document extraction, we provide each method with the same tensile and compressive strengths. Because the assembly load is transferred through a circular bore, the thick cylinder approximation provides a simple analytical baseline that does not require an FE response library. We compare both methods with refined FE solutions to determine whether the response model better captures the finite pressure patch and boundary conditions. Capacity error is measured by the absolute relative error
$e_C=\left|\widehat{C}_{\mathrm{ax}}/C^\star-1\right|\times100\%$.

Table~\ref{tab:capacity_setup} summarizes the response library and the numerical checks. We first verify the FE solver against the Lam\'e solution under uniform pressure. Both stress and displacement errors decrease as the mesh is refined. For each evaluation geometry, we compare capacities from two mesh resolutions and use the fine-mesh result as $C^\star$ only when their difference is below 3\%. All geometries satisfy this criterion, supporting the use of the refined FE capacities as numerical references.

\begin{table}[t]
\centering
\footnotesize
\setlength{\tabcolsep}{3pt}
\renewcommand{\arraystretch}{1.08}
\caption{Capacity evaluation configuration and reference checks.}
\label{tab:capacity_setup}
\begin{tabular}{@{}p{0.27\columnwidth}p{0.67\columnwidth}@{}}
\toprule
Item & Configuration or result \\
\midrule
Fixed geometry &
Bore radius $a=5$~mm, height $H=30$~mm,
loaded span centered at $z=18$~mm,
Poisson's ratio $0.2$. \\

Lam\'e verification &
Radial resolutions: 8, 16, and 32 elements.
Maximum stress error: 5.62\% to 1.55\%;
displacement error: 0.0827\% to 0.00521\%. \\

Reference meshes &
$32\times192$ and $64\times384$.
Maximum capacity change: 0.23\% on the initial
geometries and 2.19\% on the additional geometries. \\

Response library &
27 unit-pressure solutions over
$t_{\mathrm{w}}\in\{1.5,3,6\}$~mm,
$w\in\{6,12,18\}$~mm, and
$\mu\in\{0.3,0.6,0.9\}$. \\

\bottomrule
\end{tabular}
\end{table}

No tested geometry is included in the response library, and the library remains fixed throughout. Fig.~\ref{fig:capacity} reports the initial geometries, the Sobol geometries evaluated without refitting, and brick MU15, which tests transfer beyond AAC.

Figure~\ref{fig:capacity} shows that the response model of Eq.~\eqref{eq:estimated_capacity} has a mean error of 10.74\% on the initial geometries, compared with 36.54\% for the thick cylinder approximation. Without refitting, the respective mean errors are 8.00\% and 39.11\% on the additional geometries, with maximum errors of 18.20\% and 61.59\%. At a fixed geometry the four AAC grades share the unit-load stress field and the tensile criterion governs, so their relative errors coincide and panels (b) and (c) report one error value per geometry.

Reducing the response library to its eight corner points raises the initial mean error to 27.37\%, showing that the intermediate points improve interpolation. Shifting the pressure center by $\pm3$~mm gives a mean error of 10.45\% over sixteen additional stress fields. Response evaluation is about $54{,}100\times$ faster than a refined FE solve (Table~\ref{tab:llm}). These comparisons evaluate the response approximation under the prescribed continuum model; they do not independently validate the material failure criterion for a physical assembly.

\subsection{Material Evidence and Capacity Estimation}
\label{sec:evidence_capacity}
\begin{table*}[t]
\centering
\footnotesize
\caption{Capacity estimation over 80 records from 40 material--geometry conditions. Bold marks the best result in the primary group, including ties.}
\label{tab:llm}
\begin{tabular}{lrrrrr}
\toprule
Method & Valid $\uparrow$ & Mean (\%) $\downarrow$ & Median (\%) $\downarrow$ & Max (\%) $\downarrow$ & Within 20\% $\uparrow$ \\
\midrule
\multicolumn{6}{l}{\textit{Primary estimators}} \\
Identity prior, direct LLM & \textbf{80} & 79.95 & 58.78 & 2077.36 & 10 \\
Identity prior + response & \textbf{80} & 41.21 & 44.55 & 95.86 & 17 \\
Linked evidence, direct LLM & \textbf{80} & 80.65 & 41.12 & 1534.96 & 19 \\
Linked evidence + calculator & \textbf{80} & 90.47 & 41.13 & 2326.72 & 19 \\
Linked evidence + FE tool & 78 & 81.86 & 31.87 & 2345.71 & 33 \\
Basic extraction + response & 56 & \textbf{10.74} & \textbf{10.00} & \textbf{14.01} & 56 \\
Linked extraction + thick cylinder & \textbf{80} & 36.54 & 36.66 & 53.37 & 20 \\
\textbf{SAGE (best complete)}: linked extraction + response & \textbf{80} & \textbf{10.74} & \textbf{10.00} & \textbf{14.01} & \textbf{80} \\
\midrule
\multicolumn{6}{l}{\textit{Oracle and reference controls}} \\
Oracle evidence, direct LLM & 80 & 817.69 & 41.16 & 58478.22 & 18 \\
Oracle extraction + response & 80 & 10.74 & 10.00 & 14.01 & 80 \\
Linked extraction + reference FE & 80 & 0.00 & 0.00 & 0.00 & 80 \\
\bottomrule
\end{tabular}
\vspace{2pt}

{\footnotesize Response evaluation: 41.68~$\mu$s; reference FE solve: 2.255~s ($\approx54{,}100\times$ slower). Both exclude retrieval and LLM inference.}
\end{table*}

\begin{figure*}[t]
\centering
\includegraphics[width=\textwidth]{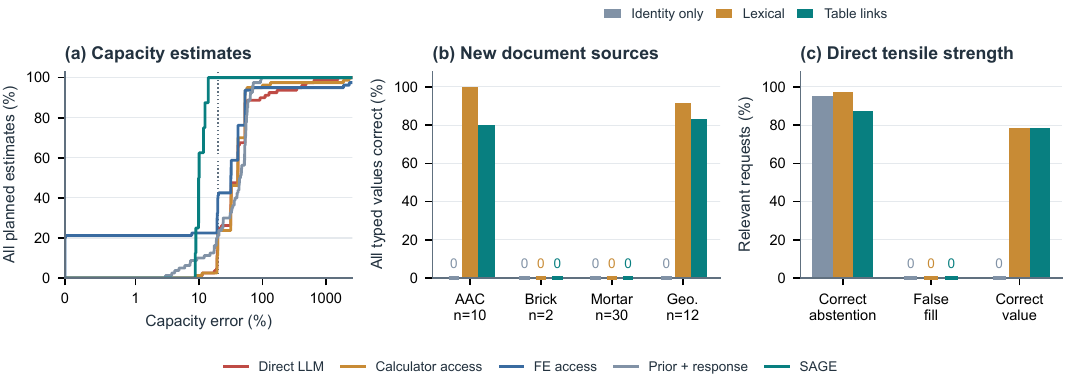}
\caption{Language model diagnostics. (a) Capacity-error coverage; the dotted line marks 20\%. (b) Complete strength recovery from four additional sources. (c) Recovery and abstention for direct tensile strength.}
\label{fig:llm}
\end{figure*}

\begin{figure*}[t]
\centering
\includegraphics[width=0.9\textwidth]{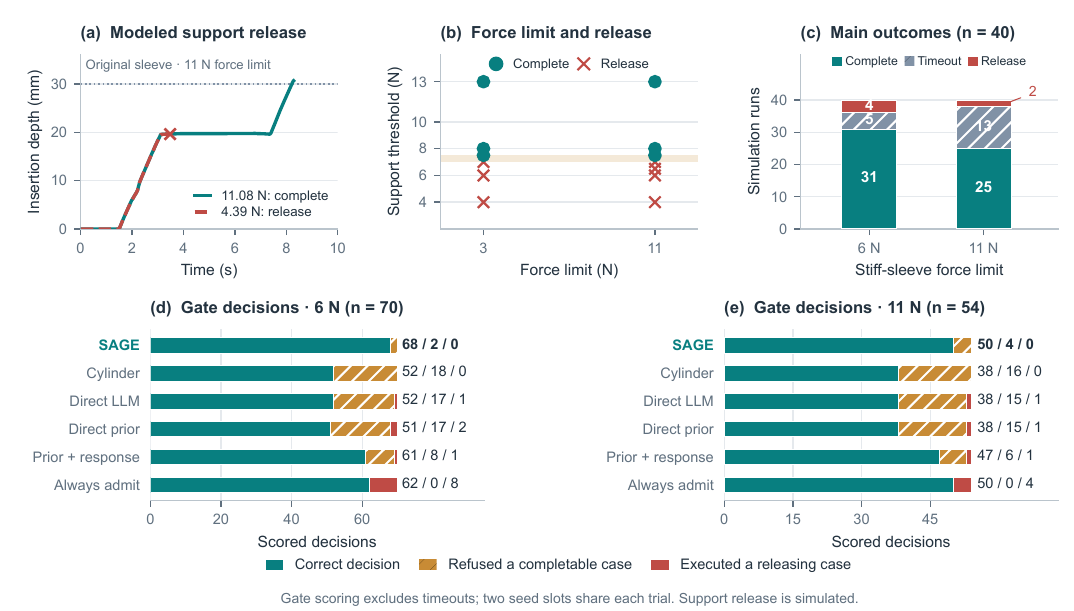}
\caption{Admission under the assigned support release model. (a) Original sleeve traces at an 11~N force limit. (b) Original sleeve support ladder; 7.0~N releases and 7.5~N holds at 3 and 11~N. (c) Outcomes for 40 stiff-sleeve cases. (d--e) Admission decisions at $B=15$~N, reported as correct / conservative refusal / admitted release. Two LLM seeds share each simulation run; timeouts are reported separately.}
\label{fig:gate}
\end{figure*}
After evaluating mechanical conversion with known strengths, we next examine whether the document stage recovers the required properties and how the resulting evidence affects capacity estimation. All language model conditions use Qwen3-32B~\cite{qwen3} with the same decoding configuration and two generation seeds.

For each method, five materials and two seeds produce ten property extractions, each reused across eight geometries, yielding 80 capacity records over 40 material--geometry conditions, as the two seeds extract identically here. In Table~\ref{tab:llm}, Basic extraction uses lexical retrieval, Linked extraction additionally follows table references, and Oracle extraction receives the relevant tables directly; SAGE combines linked extraction with the response model.

Each extraction method is paired with a conversion method and applied to matched evidence, with further conditions giving the LLM calculator or FE access. Coverage metrics include every planned case, whereas error statistics use valid numerical outputs, and missing or invalid outputs are not regenerated.

SAGE is the best complete estimator in the primary group: every case yields a valid estimate, all within 20\% of the reference, as Fig.~\ref{fig:llm}(a) shows. Its mean error matches the known-strength result of Sec.~\ref{sec:capacity_accuracy} because the relative error of Eq.~\eqref{eq:estimated_capacity} depends on the interpolated coefficients alone, not on the strength values; exact recovery therefore reproduces that error rather than reusing the same records. Direct LLM prediction from the same evidence has a mean error of 80.65\%, even though it reports both source strengths correctly. This matched comparison shows that its main error arises when converting material properties into assembly capacity. All SAGE errors are underestimates; this empirical result does not constitute a general safety guarantee.

Only the reference FE control is more accurate, because it defines $C^\star$ and therefore has zero error by construction, at the solve cost reported in Table~\ref{tab:llm}. Oracle extraction with the response model matches SAGE, so linked extraction already preserves the conversion accuracy.

Following table references increases valid capacity coverage by 30 percentage points over basic extraction without changing the conditional capacity error: linked retrieval supplies evidence coverage, while the response model supplies conversion accuracy.

Providing better evidence or optional tools does not resolve the conversion problem reliably. Correct FE-tool calls are nearly exact, but inconsistent tool use leaves the complete FE-access condition substantially less accurate than SAGE, and calculator access provides no improvement. With oracle tables the median is essentially unchanged and the inflated mean comes from a few predictions wrong by orders of magnitude, the worst returning $7682$~N against a $13.1$~N reference although both strengths were quoted correctly. These results show that access to evidence or to a solver is insufficient unless the conversion procedure is applied consistently.

A separate diagnostic extends the extraction evaluation to four additional document sources~\cite{devi2022rate,yellaiah2014tensile,rotunno2023bonding,kubica2022mortar}. Under a matched evidence budget, Fig.~\ref{fig:llm}(b)--(c) compares material identity, lexical retrieval, and retrieval with table links. Identity alone cannot recover the requested documented properties, whereas retrieval recovers reported values and generally abstains when a value is absent. Table links do not improve complete recovery on these additional sources, indicating that extraction performance remains sensitive to document and table structure.

\subsection{Force Limit Transfer to the Insertion Policy}
\label{sec:force_transfer}

After evaluating capacity estimation, we next verify its transfer to the pretrained insertion policy. Equation~\eqref{eq:force_selection} converts $\widehat{C}_{\mathrm{ax}}$ into the policy input $F_{\max}$. The policy weights remain fixed, and the estimate affects execution only through this input.

We evaluate two AAC B04 cases with refined FE capacities of $8.87$ and $10.81$~N and SAGE estimates of $7.82$ and $9.74$~N. Both estimates lie within the clip range of $3$--$11$~N and are transferred without clipping. Each force limit is tested from four fixed offsets, $(\pm0.125,0)$ and $(0,\pm0.125)$~mm, using the original sleeve. The two generation seeds produce the same limits, giving eight unique simulation runs.

All eight runs reach the target depth and pass the trajectory audit, which requires the assembly to stay within predefined position and orientation tolerances of its recorded fixture anchor throughout the run. Peak axial loads range from $7.55$ to $7.62$~N, below both the transferred $F_{\max}$ and the corresponding reference capacity. This verifies the implemented transfer and successful execution for the tested insertions with positive clearance. It is an interface check on two cases, not a comparison against fixed force limits.

\subsection{Real-Robot Experiments}
\label{sec:hardware}

We also evaluate the positive clearance pipeline on an xArm6 equipped with a wrist force/torque sensor. A construction nail serves as the peg for acrylic and foam board assemblies. Because the prior policy~\cite{priorpolicy} was trained on a different arm, it is retrained for the xArm6 with the same method and then held fixed during the recorded experiments. Each run begins from the same calibrated pose, and the PDDL interface coordinates pickup, alignment, and insertion. The axial force limit for each assembly is the capacity SAGE estimates from that material's documents and transfers through Eq.~\eqref{eq:force_selection}, as in simulation.

Before each run, the sensor is tared with 30 free-space samples, and we record the resultant force and the distance to the calibrated hole root at about 59~Hz. A run succeeds when the nail stays within 12~mm of the root for six control steps and holds its final pose; an independent hardware safety stop triggers above 15~N, beyond the policy force limit. Across both assemblies, 9 of 13 trials succeeded, and recorded videos show successful insertion. Because logging was incomplete for part of the campaign, we report one successful foam board trial with a complete log: the nail reaches 9.0~mm from the root and remains seated with a peak force of 9.6~N. Figure~\ref{fig:method} shows both hardware configurations. These runs are a feasibility demonstration, not a quantitative hardware evaluation.

\subsection{Additional Admission Check for Interference Fits}
\label{sec:admission_results}

We next evaluate whether estimated capacity can screen interference fits before execution, applying the gate of Eq.~\eqref{eq:gate} so that execution is admitted only when $\widehat{C}_{\mathrm{ax}}\ge B$.

Fig.~\ref{fig:gate}(a) shows original-sleeve insertion traces at an $11$~N force limit. Sweeping $T$ at force limits of $3$ and $11$~N, the ladder in Fig.~\ref{fig:gate}(b) gives $B=7.5$~N for the original sleeve, and the same procedure gives $B=15$~N for the stiff sleeve. We evaluate the gate on 40 material and geometry cases using the stiff sleeve. Each case is executed at two prescribed axial force limits, $6$ and $11$~N, regardless of the predicted decision, with its refined FE capacity $C^\star$ assigned as the release threshold. Fig.~\ref{fig:gate}(c) reports completion, release, and timeout separately.

Of the 80 runs, 18 time out and are not scored, leaving 35 scored runs at $6$~N and 27 at $11$~N. SAGE classifies 59 of the 62 correctly and admits none of the six that release the support; its three errors are conservative refusals, whereas always admitting accepts every releasing case. The two seeds decide identically on every run, so this equals the 118 of 124 seed-level records in Fig.~\ref{fig:gate}(d)--(e).

These results measure admission under the assigned release model of Sec.~\ref{sec:admission}, not material failure in a real interference fit.

\section{CONCLUSION}
SAGE connects material documents to robotic insertion by using an LLM to extract material strengths with their supporting source passages and a fixed mechanical response model to convert those strengths into assembly capacities. This division reduces the mean capacity error from 80.65\% for direct LLM estimates to 10.74\%, while maintaining accuracy on additional geometries without refitting. In simulation, the estimated capacities guide force limit selection for positive-clearance fits and admission for interference fits. Under the assigned support release model, the gate correctly classifies 59 of 62 scored simulation runs without admitting any execution that releases the support. Recorded xArm6 demonstrations also show that SAGE can use material documents to complete physical insertions in acrylic and foam board. Overall, SAGE provides a traceable connection from documented material evidence to mechanically grounded robotic insertion decisions.

\bibliographystyle{IEEEtran}
\bibliography{references}

\end{document}